\RequirePackage{fix-cm}
\documentclass[12pt]{article}

\usepackage{amsmath}
\usepackage{graphicx}
\usepackage{natbib}
\usepackage{url} 

\newcommand{\blind}{0}

\usepackage{epstopdf}
\usepackage{amssymb}
\usepackage[linesnumbered,algoruled,boxed,lined]{algorithm2e}
\usepackage{dsfont}
\usepackage{booktabs}
\usepackage{tikz}
\usepackage{listings}
\usepackage{color, colortbl} 
\definecolor{mygreen}{RGB}{28,172,0} 
\definecolor{mylilas}{RGB}{170,55,241}
\definecolor{Gray}{gray}{0.9}

\DeclareMathOperator*{\argmax}{argmax}
\usepackage{amsthm}
\newtheorem{proposition}{Proposition}
\newtheorem{theorem}{Theorem}
\theoremstyle{definition}
\newtheorem{definition}{Definition}
\newtheorem{remark}{Remark}

\begin{document}	
	
	\def\spacingset#1{\renewcommand{\baselinestretch}%
		{#1}\small\normalsize} \spacingset{1}

	
	\if0\blind
	{
		\title{\bf SeqROCTM: A Matlab toolbox for the analysis of Sequence of Random Objects driven by Context Tree Models}
		
		\author{Noslen Hern\'{a}ndez \hspace{.2cm}\\
			Department of Statistics, University of S\~ao Paulo\\
			and \\
			Aline Duarte \thanks{This work is part of University of S\~ao Paulo project \textit{Mathematics, computation, language and the brain}, FAPESP project \textit{Research, Innovation and Dissemination Center for Neuromathematics} (grant 2013/07699-0). Author N. Hern\'{a}ndez was fully supported by FAPESP fellowship 2016/22053-7. }\\
			Department of Statistics, University of S\~ao Paulo}
		\maketitle
	} \fi
	
	\if1\blind
	{
		\bigskip
		\bigskip
		\bigskip
		\begin{center}
			{\LARGE\bf SeqROCTM: A Matlab toolbox for the analysis of Sequence of Random Objects driven by Context Tree Models}
		\end{center}
		\medskip
	} \fi
	
	\bigskip
	\begin{abstract}
		In several research problems we deal with probabilistic sequences of inputs (e.g., sequence of stimuli) from which an agent generates a corresponding sequence of responses and it is of interest to model the relation between them. A new class of stochastic processes, namely \textit{sequences of random objects driven by context tree models}, has been introduced to model such relation in the context of auditory statistical learning. This paper introduces a freely available Matlab toolbox (SeqROCTM) that implements this new class of stochastic processes and three model selection procedures to make inference on it. Besides, due to the close relation of the new mathematical framework with context tree models, the toolbox also implements several existing model selection algorithms for context tree models.
	\end{abstract}
	
	\noindent%
	\vfill
	
		\lstset{
		language=Matlab,
		aboveskip=3mm,
		belowskip=3mm,
		showstringspaces=false,
		columns=flexible,
		basicstyle={\small\ttfamily},
		numbers=none,
		numberstyle={\tiny \color{black}},
		keywordstyle=\color{blue},
		commentstyle=\color{mygreen},
		stringstyle=\color{mylilas},
		breaklines=true,
		morekeywords={matlab2tikz},
		morekeywords=[2]{1}, keywordstyle=[2]{\color{black}},
		identifierstyle=\color{black},%
		numbersep=9pt, 
		emph=[1]{for,end,break},emphstyle=[1]\color{blue}, 
		breaklines=true,
		breakatwhitespace=true,
		tabsize=3    
	}
	
	\newpage
	\spacingset{1.5} 
\section{Introduction}
\label{sec:intro}
In several research problems we deal with probabilistic sequences of inputs (e.g., sequence of stimuli) from which an agent generates a corresponding sequence of responses and it is of interest to model or find some kind of relation between them. This is the case, for example, of many experiments related to the study of statistical learning in neuroscience. Statistical learning in this context refers to the ability to extract statistical regularities from the environment over time \citep{Armstrong_2017,CONWAY2020279,SCHAPIRO2015501}. In this kind of experiments, humans or animals are exposed to sequences of stimuli with some statistical regularity and it is conjectured that the brain is able to retrieve/learn the statistical regularities encoded in the stimuli \citep{VonHelmholtz:67,Wacongne:12,Garrido:13}. The study of this conjecture usually involves data analysis of some physiological or behavioral responses recorded from participants during the performance of a suitable task. Statistical learning is also a widely used term in computer science but it is not that meaning we are referring to here.    

Motivated by an auditory statistical learning problem, a new class of stochastic process was introduced in \cite{Duarte:19} to model the relation between the probabilistic sequence of inputs and the corresponding sequence of responses, namely \textit{sequences of random objects driven by context tree models}. A process in this class has two elements: a stochastic source generating the sequence of stimuli and a sequence of responses generated by the agent during the exposure to the stimuli. The source generating the stimuli is assumed to be a \textit{context tree model} (CTM) taking values in a categorical set \citep{Rissanen:83,buhlmann99,Galves-Loch:08}. In a context tree model, at each step, the occurrence of the next symbol is solely determined by a variable-length sequence of past symbols, called \textit{context}. Any stationary stochastic chain can be approximate by a context tree model, therefore it constitutes a powerful tool to model the probabilistic structure of sequences of stimuli \citep{FernandezGalves02,DGG06}. The model also assumes the existence of a family of probability measures on the set of responses (hereafter called objects) indexed by the contexts characterizing the sequence of stimuli. This family of measures describes the relation between the source and the responses. More precisely, at each step, a new object of the response sequence is chosen according to the probability measure associated to the context ending at that step in the sequence of stimuli.  

Given some data, statistical model selection methods can be applied to estimate the set of contexts and, in some cases, the family of probability measures characterizing the distribution of the response sequence. The theoretical framework introduced in \cite{Duarte:19} addressed a model selection algorithm for the particular case in which the objects of the response sequence refer to functions (e.g., electrophysiological data). Nevertheless, the proposed mathematical framework is general enough to model other kind of objects in the response sequences such as responses taking values in a categorical set, in the real line or in $\mathds{R}^n$. In the same way that context tree models have found applications in several research areas such as neuroscience, genetics \citep{Leonardi2009} and linguistics \citep{galves:12}, this new theoretical framework can find applications in research areas beyond neuroscience. For this reason, its computational implementation is useful. 

This paper introduces the SeqROCTM Matlab toolbox, a set of computational tools for working with sequences of random objects driven by context tree models. In addition to the statistical software, this article makes other contributions. We formalize two model selection procedures for sequences of random objects driven by context tree models with categorical responses. We also formalize and  define general conditions for the use of the Smaller Maximizer Criteria (SMC) to tune model selection algorithms for both context tree models and sequence of random objects driven by context tree models. 
 
Since the stimuli sequence is generated by context tree model, the SeqROCTM toolbox also implements  several model selection algorithms and tuning procedures for context tree model. Up to our knowledge, there exist only an R package implementing model selection for context tree model \citep{Machler2004}. 

The paper is organized as follows. Section \ref{sec:modelselection} briefly describes the mathematical concepts and model selection algorithms included in the SeqROCTM toolbox. Section \ref{sec:soft_framework} introduces the architecture of the software. Sections \ref{sec:example1} and \ref{sec:example2} present the main functionalities of the toolbox through illustrative examples motivated by statistical learning problems. Conclusions are given in Section \ref{sec:conclusions}.

\section{Model selection methods for sequence of random objects driven by context tree models}
\label{sec:modelselection}

We begin this section by introducing some notation and formal definitions. Let $A$ be a finite set. Any string $u = (u_{n-m},...,u_{n-1}) \in A^m$, $1\leq m\leq n,$ is denoted by $u_{n-m}^{n-1}$ and its length by $l(u)$.
Given two strings $u$ and $v$ of elements of $A$, $uv$ denotes the string of length $l(u)+l(v)$ obtained by concatenating $u$ and $v$. The string $u$ is said to be a \textit{suffix} of $v$, and denote by $u\preceq v$, if there exists a string $s$ satisfying $v = su$. When $v \neq u$ we say that $u$ is a \textit{proper suffix} of $v$ and denote $u\prec v$. 

\begin{definition}
A \textit{context tree} is defined as any set $\tau\subset A^* = \cup_{m= 1}^\infty A^m$ satisfying
\begin{enumerate}
\item \textit{Suffix Property}. No string $w \in \tau$ is a proper suffix of another string  $s\in\tau$.
\item \textit{Irreducibility}. No string belonging to $\tau$ can be replaced by a proper suffix without violating the suffix property.
\end{enumerate}
\end{definition}

The elements of $\tau$ are called \textit{contexts}. The \textit{height} and \textit{size} of $\tau$ are defined as $l(\tau) = \max\{l(w): w \in \tau\}$ and $|\tau|$, respectively. 

In this work we deal with experiments in which an agent is exposed to an stochastic sequence of stimulus $(X_n)_n$, taking values in $A$. A key point is that the stochastic sequence $(X_n)_n$ is a context tree model \citep{Rissanen:83,buhlmann99}. Formally, consider a stationary ergodic process $(X_n)_{n\geq 1}, \, X_n\in A,$ and for any string $s\in A^*$ denote by 
\[
p(s)=P\big(X^{l(s)}_1=s\big).
\]

\begin{definition} We say  $(X_n)_n$ is a \textit{context tree model} with parameters $(\tau, p)$ if there exist a function $c_{\tau}:A^*\to \tau$ such that
\begin{enumerate}
	\item for any $n \geq l(\tau)$ and any finite sequence $x_{-n}^{-1} \in A^n$ such that $p(x_{-n}^{-1})>0$, it holds that
	$$P\big(X_{n+1} = a | X_{1}^{n}= x_{-n}^{-1}\big) = p\big( X_{n+1}=a | c_{\tau}(x_{-n}^{-1})\big) \mbox{ for all } a \in A. $$
	\item no proper suffix of $c_{\tau}(x_{-n}^{-1})$ satisfies condition 2.
\end{enumerate}
The function $c_{\tau}$ is said a \textit{context function}.
\end{definition}

A sequence of responses $(Y_n)_n$ with values in some measurable space $(F,\mathcal{F})$ is recorded while the agent is exposed to the stimuli sequence (e.g., neurophysiology responses such as electroencephalographic data, behavioral responses, etc.). The relation between the responses and the stimuli is modeled through a  class of stochastic processes called sequence of random objects driven by context tree model \citep{Duarte:19}. 

\begin{definition} The bivariate stochastic chain $(X_{n},Y_{n})_n$ taking values in $A \times F$ is a \textit{sequence of random objects driven by context tree models} with parameters $(\tau, p, q)$, where $q$ is a family of probability measures on $(F,\mathcal{F})$, if
\begin{enumerate}
	\item $(X_{n})_n $ is a context tree model with parameters $(\tau,p)$;
	\item conditionally to the sequence $\left( X_{n}\right)_n$, $(Y_n)_n$ are independent random variables  and, for any $n \geq l(\tau)$, it holds that
$$
P\left(Y_{n}\in J | X_{1}^{n} = x_{1}^{n} \right) = q\left( Y_{n}\in J | c_{\tau}\left(x_{1}^{n}\right) \right) \mbox{ for any } \mathcal{F}\mbox{-measurable } J.
$$  
\end{enumerate}
\end{definition}

Given some training data $(X_n,Y_n)_n$ and, under the assumption that it was generated by a sequence of random objects driven by context tree models, the statistical problem of interest is to estimate the parameters $\tau$ and $q$  characterizing the response sequence. 

To formulate the model selection methods we consider two scenarios. A first scenario in which the responses $Y_n$ belong to a finite set and a second scenario in which $Y_n$ takes values in a functional space. Hereafter we will refer to these two scenarios as \textit{categorical} and \textit{functional} case, respectively.

Before introducing the model selection procedures a few more definitions are needed. Given a finite string $u \in A^*$  we denote by $N_n^X(u)$ the number of occurrences of the string $u$ in the sequence $(X_1,...,X_{n-1})$, that is
\begin{equation}
N_n^X(u) = \sum_{t = l(u)}^{n-1}\mathds{1}_{\{X_{t - l(u)+1}^t = u\}}.
\end{equation}

\begin{definition} 
Let $L$ be an integer such that $1 \leq L \leq n$, an \textit{admissible context tree of maximum height $L$} for the sample $(X_1,...,X_n)$ is any context tree $\tau$ satisfying 
\begin{itemize}
\item[i)] $w \in \tau$ if and only if $l(w) \leq L$ and $N_n^X(w) \geq 1$.
\item[ii)]  Any string $v \in A^*$ with $N_n^X(v) \geq 1$ is a suffix of some $w \in \tau$ or has a suffix $w \in \tau$.
\end{itemize}
The set of all admissible context trees of maximal height $L$ is denoted by $\Gamma^L(X^n_1)$.
\end{definition}

\begin{definition}
	Let $\tau$ be a context tree and fix a finite string $s \in A^*$. A \textit{subtree} in $\tau$ induced by $s$ is defined as the set $\tau_s=\{w \in \tau : s\prec w \}$. The set $\tau_s$ is called a \textit{terminal subtree} if for all $w \in \tau_s$ it holds that $w = as$ for some $a \in A$. 
\end{definition}

Given two context trees $\tau_1$ and $\tau_2$, we denote by $\tau_1\preceq \tau_2$ (resp. $\prec$) if for any $w\in \tau_1$ there exists $s\in \tau_2$ such that $w\preceq s$ (resp. $\prec$).

The model selection procedure that will be introduced for the functional case and two out of three procedures developed for the categorical case are inspired by the algorithm Context \citep{Rissanen:83}. For this reason, we first describe below a general algorithm Context procedure, specifying later the difference on each case. In the sequence, we present a third model selection procedure for the categorical case which is based on BIC.

\subsection{General algorithm Context}

Given a sample $(X_1,Y_1),...,(X_n,Y_n)$, fix an integer $1 \leq L \leq n$ and let $\mathcal{T}_n^L(X^n_1)$ be the context tree of maximum size in $\Gamma^L(X^1_n)$. We shorten this maximal candidate context tree $\hat{\tau} = \mathcal{T}_n^L(X^n_1)$ by successively pruning the terminal subtrees according to some statistical criterion. 

For any string $u \in A^*$ such that $\hat{\tau}_u$ is a terminal subtree, we decide to prune or not $\hat{\tau}_u$ verifying a statistical criterion, say \texttt{stat(u)}. If \texttt{stat(u)} is satisfied, we prune the subtree $\hat{\tau}_u$ in $\hat{\tau}$,
\begin{equation}
\hat{\tau} = \left( \hat{\tau} \setminus \hat{\tau}_u \right) \cup \{u\}.
\end{equation}
Otherwise, if \texttt{stat(u)} is not satisfied, we keep $\hat{\tau}_u$ in $\hat{\tau}$. At each pruning step, we check a string $s \in A^*$ which induces a terminal subtree in $\hat{\tau}$ and that has not been checked yet. This pruning procedure is repeated until all the existing terminal subtrees have been checked and no more pruning is possible.
 
A pseudo code for the general algorithm Context is given in Algorithm \ref{alg:context_tree_estimation}.

\begin{algorithm}[H]
	\KwIn{An alphabet $A$, a sample $(x_1,y_1),...,(x_n,y_n)$  with $x_k \in A, y_k \in \mathcal{Y}$ for $1 \leq k \leq n$, a positive integer $L$.}
	\KwOut{A context tree $\hat{\tau}$ and a family of distributions $\hat{q}$ (only in the categorical case) indexed by the elements of $\hat{\tau}$ .}   
	
	Compute $\mathcal{T}_n^L(x_1^n)$ and initialize $\hat{\tau} \gets \mathcal{T}_n^L(x_1^n)$
	
	Flag$(s) \gets $ ``not checked'' for all string $s$ such that $s \prec w \in \hat{\tau}$
	
	\While{$\exists s \in A^*$: $\hat{\tau}_s$ is a terminal subtree and Flag$(s) = $ ``not checked''}
	{Compute the statistic criterion \texttt{stat(s)} 
		
		\uIf{\texttt{stat(s)} is satisfied} 
		{
			$\hat{\tau} \gets (\hat{\tau} \setminus \hat{\tau}_s) \cup \{s\}$  
		}
		\Else
		{
			Flag$(s) \gets $ ``checked''
		}
	}
	Compute $\hat{q}$
	
	\Return{$\hat{\tau}, \hat{q}$}
	
	\caption{General algorithm Context for sequences of random objects driven by context tree models}
\label{alg:context_tree_estimation}
\end{algorithm}

The use of the general algorithm Context for the functional and categorical cases differs with respect to \texttt{stat(u)}. 

In the functional case \texttt{stat(u)} implements a two-side goodness of fit test for functional data \citep{Cuesta2006}. For the categorical case, \texttt{stat(u)} implements two different procedures. The first one compares the conditional log-likelihoods between a string and its offspring while the second one compares the empirical distributions between a string and its offspring. The following sections describe the statistical criterion \texttt{stat(u)} implemented in each case.
 
\subsection{Functional case}

In this section it is defined the statistic criterion used in the general algorithm Context  (Algorithm \ref{alg:context_tree_estimation}) for the functional case.

Consider $F = L^2([0,T])$ the set of real-valued square integrable function in the interval $[0,T]$ and $\mathcal{F}$ the Borel $\sigma-$algebra on $L^2([0,T])$. Moreover, assume that $(X_1,Y_1),\ldots, (X_n,Y_n)$ is a sample of a sequence of random objects driven by context tree models with parameters $(\tau^*,p^*,q^*)$, with $q^*$ a probability measure on $(F,\mathcal{F}).$
For any string $s \in A^*$ with $l(s) \leq L$, let $I_n(s)$ be the set of indexes belonging to $\{l(s),...,n\}$ where the string $s$ occurs in the sample $(X_1,...,X_n)$, that is
\begin{equation}
	I_n(s) = \{l(s) \leq m \leq n : X_{m-l(s)+1}^{m} = s\}.
\end{equation} 
By definition, the set $I_n(s)$ has $N_n^X(s)$ elements. If $I_n(s) = \{m_1,...,m_{N_n^X(s)}\}$, we set $Y_k^{(s)}=Y_{m_k}$ for each $1 \leq k \leq N_n^X(s)$. Thus, $Y_1^{(s)},...,Y_{N_n^X(s)}^{(s)}$ is the \textit{subsample of $(Y_1,...,Y_n)$ induced by the string} $s$. 

To each element of the terminal subtree $\hat{\tau}_u$ we associate the subsample of $(Y_1,...,Y_n)$ induced by it. These sets of functions are used to decide whether the subtree is pruned or not: the \texttt{stat(u)} criterion tests the equality of distribution between the sets of functions associated to the elements of $\hat{\tau}_u$.

To test whether two samples of functions have the same distribution, it is used a projective method proposed in \cite{Cuesta2006}. In this method, each function of both sets is projected in a gaussian direction chosen at random. This produces two new projected samples of real numbers. Then, the equality of distribution for these new samples is tested using the Kolmogorov-Smirnov test. The results presented in \cite{Cuesta2006} guarantees that, if the two samples of functions have different distributions then the set of random directions where the projected samples have the same distribution has Lebesgue measure equal to zero. This implies that if we reject the null hypothesis of equality of distributions for two sets of projected samples, then we can also reject it for the corresponding functional sets.  

Formally, for any string $u \in A^*$ such that $\hat{\tau}_u$ is a terminal subtree, we test the null hypothesis
\begin{equation}\label{eq:H0-functional}
H_0^{(u)}: \mathcal{L}\left(Y_1^{(s)}, ..., Y_{N_n^X(s)}^{(s)}\right) = \mathcal{L}\left(Y_1^{(v)}, ..., Y_{N_n^X(v)}^{(v)}\right), \,  \forall \, s, v \in \hat{\tau}_u,
\end{equation}
using the test statistic 
\begin{eqnarray}\label{eq:Delta-funcional}
\Delta_n(u) = \Delta_n^W(u) & = & \max_{s,v \in \hat{\tau}_u}D_n^W\Big((Y_1^{(s)}, ..., Y_{N_n^X(s)}^{(s)}), (Y_1^{(v)}, ..., Y_{N_n^X(v)}^{(v)})\Big) \nonumber \\
& = & \max_{s,v \in \hat{\tau}_u} \sqrt{\frac{N_n^X(s)N_n^X(v)}{N_n^X(s) + N_n^X(v)}}KS(\hat{Q}_n^{s,W},\hat{Q}_n^{v,W}).
\end{eqnarray}
Here $W$ is a realization of a Brownian bridge in the interval $[0,T] $ and $KS\big(\hat{Q}_n^{s,W},\hat{Q}_n^{v,W}\big)$ is the Kolmogorov-Smirnov distance between the empirical distributions $\hat{Q}_n^{s,W}$ and $\hat{Q}_n^{v,W}$ of the projections of the samples $Y_1^{(s)}, ..., Y_{N_n^X(s)}^{(s)}$ and $Y_1^{(v)}, ..., Y_{N_n^X(v)}^{(v)}$ onto the direction $W$, respectively.

We reject the null hypothesis $H_0^{(u)}$ when $\Delta_n(u) > \delta_{\alpha}$, with $\delta_{\alpha} = \delta_{\alpha}(u) = \sqrt{-1/2 \ln (\alpha/2M)}$ and $M ={|\hat{\tau}_u| \choose 2}$. Observe that for a string $u$ with only one pair of offspring (i.e., $|\hat{\tau}_u|=2$) the null hypothesis $H_0^{(u)}$ is tested with significance level $\alpha$ (because $\delta_\alpha$ becames the $(1-\alpha)$-percentile of a KS distribution). On the other hand, for a string $u$ with more than two offspring, $H_0^{(u)}$ is tested with an unknown significance level upper bounded by $\alpha$. This is guaranteed because in the definition of $\delta_{\alpha}$ we applied a terminal subtree correction, by using $\alpha/M$ instead of $\alpha$. In this way, $H_0^{(u)}$ is tested with significance level at most $\alpha$ for any string $u$.

The statistic in equation \eqref{eq:Delta-funcional} depends on the a random direction $W$. For this reason, to improve the stability of the estimate we test the null hypothesis \eqref{eq:H0-functional} several times using different Brownian bridges. We conclude that the sets of functions associated to the elements of the terminal subtree $\hat{\tau}_u$ do not have the same distribution, and consequently, we do not prune the subtree $\hat{\tau}_u$, if the number of rejections exceeds a certain threshold. This threshold is derived according to a binomial distribution with probability of success $\alpha$.
Formally, fixed a string $s\in A^*$, a significant level $\alpha$ and consider $N$ independent Brownian bridges $W_1,\ldots,W_N$. Compute the test statistics $\Delta^{W_1}_n, \ldots, \Delta^{W_N}_n$ and define
\begin{equation}\label{eq:DeltaBar-functional}
\bar \Delta_n(s)=\sum^N_{m=1} 1\{\Delta^{W_m}_n> \delta_{\alpha}\}.
\end{equation}
In this case, \texttt{stat(u)} checks whether $\bar \Delta_n(s) < C$,  where $C$ is the $(1-\beta)$-percentile of a binomial  distribution with parameters $N$ and $\alpha$. The elucidation for this procedure comes from the following proposition. 

\begin{proposition}\label{prop:binomial}
For any string $u\in A^*$ and integer $N\geq 1$, consider the random variables $\Delta^{W_1}_n(u), \ldots, \Delta^{W_N}_n(u)$ defined in \eqref{eq:Delta-funcional} with $W_1,\ldots,W_N$ independent Brownian bridges in the interval $[0,T]$. For any significance level $\alpha>0$ define $\delta_\alpha(u)=\sqrt{-1/2 \ln (\alpha/2M_u)}$ with $M_u={|\tau_u|\choose 2}$. Under the null hypothesis \eqref{eq:H0-functional}, for any $\beta\in (0,1)$, it holds that,
\begin{equation*}
P\left(\sum^N_{m=1} 1_{\{\Delta^{W_m}_n> \delta_\alpha (u)\} } > C_\beta \right) \leq \beta.
\end{equation*}
where $C_\beta$ denotes the smallest constant such that $P(\xi > C_\beta) \leq \beta$ with $\xi$ a random variable with Binomial distribution of parameters $N$ and $\alpha$.
\end{proposition}


\subsection{Categorical case}\label{sec:StatCriteria}

\subsubsection{General algorithm Context + Conditional log-likelihood}

Given $u \in A^*$ such that $N_n^X(u) \geq 1$, we denote by $N_n^{XY}(u,a)$ the number of occurrences of the string $u$ in the sample $(X_1,...,X_n)$ followed by the occurrence of the symbol $a$ in the sample $(Y_1,...,Y_n)$, that is 
\begin{equation}
N_n^{XY}(u,a) = \sum_{t = l(u)}^{n-1}\mathds{1}_{ \{ X_{t-l(u)+1}^t = u; Y_{t+1 } = a \} }.
\end{equation}

The maximum conditional likelihood for a sample $(X_1,Y_1),\dots. (X_n,Y_n)$ is given by
\begin{equation}\label{eq:Ltau}
 L_{(\tau,\hat q)}(Y_1^n\mid X^n_1) =\prod_{u\in \tau}  L_{(u,\hat q)}(Y_1^n\mid X^n_1)
\end{equation}
with 
\begin{equation}\label{eq:Lw}
 L_{(u,\hat q)}(Y_1^n\mid X^n_1) =\prod_{a \in A} \hat q(a|u)^{N_n^{XY}(u,a)},  
\end{equation}
and $\hat q(a|u)$ the maximum likelihood estimator of the conditional probability $q(a|u)$, defined as
\begin{equation} \label{eq:emp_transition_prob}
\hat{q}(a|u) = \frac{N_n^{XY}(u,a)}{N_n^X(u)} = \frac{N_n^{XY}(u,a)}{\sum_{a'\in A} N_n^{XY}(u, a')} .
\end{equation}

Notice that $L_{(u,q)}(X_1^n|Y_1^n)$ is the portion of the conditional likelihood $L_{(\tau,q)}(X_1^n|Y_1^n)$ of the model $(\tau, p, q)$ given the data $(X,Y)_1^n$ induced by the context $u\in \tau$.

Consider the statistic
\begin{equation}\label{eq:stat}
\Delta_n(u) = \sum_{b \in A}\sum_{a \in A} N^{XY}_n(bu,a) \log \frac{\hat{q}(a|bu)}{\hat{q}(a|u)},
\end{equation}
and fix a threshold $\delta >0$. For any string $u \in A^*$ such that $\hat{\tau}_u$ is a terminal branch, the function \texttt{stat(u)} verifies whether the following inequality holds 
\begin{equation}\label{eq:context_inequality}
\Delta_n(u) < \delta.
\end{equation} 

If the inequality is satisfied, we prune the subtree $\hat{\tau}_u$ of $\hat{\tau}$. Otherwise, we keep $\hat{\tau}_u$ in $\hat{\tau}$. The inequality (\ref{eq:context_inequality}) is equivalent to check whether $$ \sum_{b\in A} \log ( L_{(bu,\hat q)} (Y_1^n\mid X^n_1)) - \log ( L_{(u,\hat q)}(Y_1^n\mid X^n_1))  < \delta.$$

Notice that $\Delta_n(u)$ is the conditional log-likelihood ratio between a model with parameters $(\tau, p, q)$ and a model with parameters $(\tau', p, q')$,  where $\tau \succ \tau'$ and they differ only by one set of offspring nodes branching from $u$, that is $\tau'=\tau\setminus \tau_u\cup \{u\}$.

\begin{remark}
	The idea of comparing the maximum likelihood induced by a node with the maximum likelihood induced by its offspring was originally used in the algorithm Context introduced by \cite{Rissanen:83}.
\end{remark}

Given $\mathcal{T}^L_n=\mathcal{T}^L_n(X^n_1)$, set $C_w((X,Y)^n_1)=0$ for all $w\in\mathcal{T}^L_n$, and, for any $u \prec w \in \mathcal{T}^L_n$ define
\begin{equation}\label{eq:ind_ctxt}
C_{u,n} = C_u((X,Y)^n_1)=\max \big \lbrace 1_{\{\Delta_n(u)\geq \delta\}}, \max_{b\in A}C_{bu,n}) \big\rbrace.
\end{equation}
The context tree estimator $\hat{\tau}^\delta_{C,n}= \hat{\tau}^\delta_C((X,Y)^n_1)$ obtained with this procedure can be defined as
\begin{equation} \label{eq:tree_ctxt}
\hat{\tau}_{C,n}^\delta=\{w \preceq v \in \mathcal{T}^L_n: C_{w,n}=0 \mbox{ and } C_{u,n}=1 \mbox{ for all  } u\prec w\}.
\end{equation}

Notice that once we have $C_{w,n}=1$, for a given $w$, equation \eqref{eq:ind_ctxt} implies that for any $u\prec w$, $C_{u,n}=1$.

\begin{remark}\label{rmk:consistency_AC}
	The consistency of the original algorithm Context was proved in \cite{Rissanen:83}. In this setting (i.e., model selection procedure for context tree model) the statistic used to identify the contexts is given by
	\begin{equation*}
	\Delta^X_n(u) = \sum_{b \in A}\sum_{a \in A} N^{X}_n(bua) \log \frac{\hat{p}(a|bu)}{\hat{p}(a|u)}.
	\end{equation*}
	The proof of consistency depends on $\hat p$ through the ergodicity of $p$ and its memory relation with $\tau$, which in our case $q$ also satisfies. Therefore, the consistency can be easily adapted for the formulation we are introducing here for sequences of random objects driven by context tree models.
\end{remark}

\subsubsection{General algorithm Context + Offspring empirical distributions}

In this case, the function \texttt{stat(u)} inside the general algorithm Context compares the distance between the empirical distribution associate to the string $u$ and the ones associated to its offspring.   

Formally, for any finite string $u \in A^*$, define the statistic
\begin{equation}
	\tilde{\Delta}_n(u) =  \max_{b \in A} \Big(\max_{a \in A} |\hat{q}(a|u) - \hat{q}(a|bu)| \Big).
\end{equation}

For any finite string $u \in A^*$ such that $\tau_u$ is a terminal subtree, the function \texttt{stat(u)} verifies whether $\tilde{\Delta}_n(u) < \delta$. If the inequality is satisfied the subtree $\tau_u$ is pruned, otherwise it is kept.

Given $\mathcal{T}^L_n=\mathcal{T}^L_n(X^n_1)$, set $\tilde C_w((X,Y)^n_1) = 0$ for all $w\in\mathcal{T}^L_n$, and, for any $u \prec w \in \mathcal{T}^L_n$ define
\begin{equation}\label{eq:ind_dtxt}
\tilde C_{u,n} =\max\{  1_{\{\tilde{\Delta}_n(u)\geq \delta\} },\, \max_{b\in A}\{ C_{bu,n} \}   \}.
\end{equation}

The context tree estimator $\hat{\tau}^\delta_{\tilde C,n}= \hat{\tau}^\delta_{\tilde C}((X,Y)^n_1)$ obtained with this procedure can be defined as 
\begin{equation} \label{eq:tree_dtxt}
\hat{\tau}_{\tilde C,n}^\delta=\{w \preceq v \in \mathcal{T}^L_n: \tilde C_{w,n}=0 \mbox{ and }\tilde C_{u,n}=1 \mbox{ for all  } u\prec w\}.
\end{equation}
where $\mbox{suf}(w)$ refers to the largest suffix of $w$. Note that, as well as in the classical algorithm Context, $\tilde C_{u,n} = 1$ implies $\tilde C_{v,n} = 1$ for all $v \prec u$.

\begin{remark}\label{rm:consis_mod_AC}
In \cite{Galves-Leonardi2008} it was proved the strong consistency of the estimator \eqref{eq:tree_dtxt} (with $\hat p$ instead of $\hat q$) for the case of unbounded context tree models. The proof relies on a mixture property which is always satisfied in the case of finite context tree models. In particular, is also true for the law of the response sequence $(Y_n)_n$ since its time memory depends on the time memory of the associated context tree model.
\end{remark}

\subsubsection{Bayesian Information Criterion (BIC)}

This section describes a model selection procedure for the categorical case using the Bayesian Information Criterion. Model selection for context tree models via BIC was first addressed in \cite{CsiszarTalata06}. We formalize here how the procedure introduced in \cite{CsiszarTalata06} is used in our case.

Given a sample $(X,Y)^n_1$ and a constant $c>0$, the BIC estimator for sequence of random objects driven by context tree models is defined as 
\begin{equation}\label{eq:est_BIC}
\hat{\tau}_{BIC,n}^c=\hat{\tau}^c_{BIC}((X,Y)^n_1)= \argmax_{\tau \in \Gamma^L_n}\Big\lbrace 
\log  L_{(\tau,\hat q)} - c\cdot df(\tau)\log(n)
\Big\rbrace.
\end{equation}
where $df$ stands for the degree of freedom of the model. Formally, for any admissible context tree $\tau$, we define, for each $w\in \tau$, 
\[
df(w)= \sum_{a\in A} 1_{\{N_n^{XY}(w,a)\geq 1\}} - 1
\] 
and  $df(\tau)=\sum_{w\in \tau} df(w)$.

\cite{CsiszarTalata06} showed that \eqref{eq:est_BIC} can be computed efficiently through the following inductive procedure. Starting with $\mathcal{T}^L_n$, for any $w\in \mathcal{T}^L_n$, define the quantity $V_{w,n} = V_w((X,Y)^n_1) = n^{-c\cdot df(w)} L_{(w,\hat q)}((X,Y)^n_1)$ and the indicator $\mathcal{X}_{w,n}=\mathcal{X}_w((X,Y)^n_1)=0$, and for any $w\prec u \in \mathcal{T}_n^L$ define recursively the quantity
\begin{equation}\label{eq:V_BIC}
V_{w,n} = \max\Big\lbrace n^{-c\cdot df(w)} L_{(w,\hat q)}((X,Y)^n_1)\, , \, \prod_{b\in A}V_{bw,n} \Big\rbrace
\end{equation}
and the indicator
\begin{equation}\label{eq:ind_BIC}
\mathcal{X}_{w,n} = 1\Big\lbrace \prod_{b\in A}V_{bw,n} >  n^{-c\cdot df(w) } L_{(w,\hat q)}((X,Y)^n_1) \Big\rbrace.
\end{equation}
The estimate obtained solving \eqref{eq:est_BIC} can be written as
\begin{equation}\label{eq:tree_BIC}
\hat{\tau}_{BIC,n}^c=\{w \preceq s \in \mathcal{T}^L_n: \mathcal{X}_{w,n}=0 \mbox{ and } \mathcal{X}_{u,n}=1 \mbox{ for all  } u\prec w\}.
\end{equation}

Observe that, on the contrary to the algorithm Context, $\mathcal{X}_{w,n}=1$ for a given $w$, does not imply that $\mathcal{X}_{u,n}=1$  for any $u\prec w$.

\begin{remark}\label{rmk:BIC_Xn}
	The fact that the recursive procedure above effectively solves the BIC optimization problem and the consistency of the estimator were proved in \cite{CsiszarTalata06} for the case of context tree models $(X_n)_n$. In \cite{CsiszarTalata06}, the analogous to equation \eqref{eq:est_BIC} is
	\begin{equation*}
	\hat{\tau}^c_{BIC}(X^n_1)=\argmax_{\tau \in \Gamma^L_n} \Big\lbrace \log  L_{(\tau,\hat p)}(X_1^n) - c\cdot df(\tau)\log(n)
	\Big\rbrace,
	\end{equation*}
in which the maximum log-likelihood is computed using  the empirical probabilities of the distribution $p$, instead of $q$. Since the distribution $p$ affects these proofs only through the ergodic theorem and its memory dependency on $\tau$,  it is straight forward that all the proofs can be adapted to the case of the conditional log-likelihood considered in this section, which depends on $\hat{q}$ instead of $\hat{p}$.
\end{remark}

\subsubsection{Tuning the model selection methods}   

The threshold $\delta$ used in the procedures based on the general algorithm Context and the penalization constant $c$ involved in the model selection procedure based on BIC are hyperparameters whose values must be specified a priori. Small values of $\delta$ and $c$ result in big context trees (big in the sense of its size) and, consequently, overfitted models while high values of these hyperparameters give rise to context trees of small size and underfitted models. 

To choose the value of the hyperparameters one can use the Smallest Maximizer Criterion (SMC) \citep{galves:12}. The SMC procedure was introduced in \cite{galves:12} to tune the model selection method for context tree models based on BIC. Here we extend this framework to the case of sequence of random objects driven by context tree models for tuning the model selection methods proposed in the categorical case.

The SMC procedure consists of two steps. In the first step a set of candidate models is computed, namely the \textit{champion trees}. In the second step, an optimal model is chosen within the set of champion trees. The champion trees obtained will depend on the model selection procedure being tuned and may differs from one procedure to another. 

In this section, $\hat{\tau}_n^\ell$ denote either $\hat{\tau}^\ell_{BIC,n}$, $\hat{\tau}^\ell_{C,n}$ or $\hat{\tau}^\ell_{\tilde C,n}$.

\begin{enumerate}
	\item[] \textbf{Step 1. Compute the champion trees.} The champion trees constitute a set of estimated context trees $\hat{\tau}^\ell_n$ obtained by varying the value of the hyperparameter $\ell\geq 0$. When $\ell = 0$ we obtain the admissible context tree of maximum size $\hat{\tau}^0_n = \mathcal{T}_n^L$ (the more complex model). By successively increasing the value of $\ell$, we obtain a finite set of context trees totally ordered with respect to the order $\succ$, say $\mathcal{C}_n = \{\mathcal{T}_n^L = \hat{\tau}_0 \succ \hat{\tau}_1 \succ ... \succ \hat{\tau}_{K} = \tau_{root}\}$. It is not hard to see that there exists a value $\ell= \ell_{max}$ such that for any $\ell \geq \ell_{max}$ the estimated model is the empty tree, $\tau_{root}=\emptyset$, which refers to the independent model. 
\end{enumerate}

A crucial fact for the consistency of SMC is that the context tree generating the sample data belongs eventually almost surely to the set of champion trees as $n$ goes to $\infty$. 
This is the content of the theorem below and its proof is a co-factor of Theorem 6 in \cite{galves:12} and an extra argument given in Appendix \ref{app:proof_orderedChTrees}.
\begin{proposition}\label{thm:tau_in_Cn}
Assume $(X_1,Y_1), \cdots, (X_n,Y_n)$ is a sample of a sequence of random objects driven by context tree model with parameters $(\tau^*, p^*, q^*)$, with $|\tau^*|\leq L$. Consider the map $\ell\in [0,+\infty]\mapsto \hat \tau^\ell_n\in \Gamma^L(X^1_n)$ with $\hat \tau^\ell_n$ denoting either $\hat \tau^\ell_C((X,Y)^n_1)$, $\hat \tau^\ell_{BIC}((X,Y)^n_1)$ or $\hat \tau^\ell_{\tilde C}((X,Y)^n_1)$ and denote by
\begin{equation}
\mathcal C_n=\{\hat \tau^\ell_n: \ell \in [0,+\infty]\}.
\end{equation} 
Then $\mathcal{C}_n$ is totally ordered with respect to $\succ$ and eventually almost surely $\tau^*\in \mathcal{C}_n$ as $n\to \infty$.
\end{proposition}

It is well known that the bigger the context tree, the higher its sample likelihood. When SMC was introduced for tuning the BIC model selection algorithm for context tree models, \cite{galves:12} theoretically proved the existence of a change of regime in the rate in which the sample likelihood increases in the set of champion trees (Theorem 7 in \cite{galves:12}). 
The authors also showed that such changing point in the likelihood function occurs at the true model generating the data. A consequence of the proof of consistency of SMC is that the change of regime does not depends on the estimation method used to obtain the champion trees, but only on some properties of the set. For this reason we state the next theorem in a slightly more general form that stated in \cite{galves:12} and in terms of sequences of random objects driven by context tree models.
\begin{theorem}\label{thm:ChangeInRegime}
Assume $(X_1,Y_1), \cdots, (X_n,Y_n)$ is a sample of a sequence of random objects driven by a context tree model with parameters $(\tau^*, p^*, q^*)$, with $|\tau^*|\leq L$. Given a set $\mathcal C_n \subset \Gamma^L(X^1_n)$ satisfying
\begin{enumerate}
\item[(i)] $\mathcal C_n$  is totally ordered with respect to $\succ$ and
\item[(ii)]  eventually almost surely $\tau^*\in \mathcal{C}_n$ as $n\to \infty$.
\end{enumerate}
The following holds:
\begin{enumerate}
\item For any $\tau\in \mathcal{C}_n,$ with $\tau\prec \tau^*,$ there exists a constant $c(\tau^*,\tau)>0$ such that
\begin{equation}
\log L_{(\tau^*,\hat q)}-\log L_{(\tau,\hat q)}\geq c(\tau^*,\tau) n 
\end{equation}
\item For any $\tau\prec \tau'\in \mathcal{C}_n,$ with $\tau^*\preceq \tau,$ there exists a constant $c(\tau',\tau)>0$ such that
\begin{equation}
\log L_{(\tau',\hat q)}-\log L_{(\tau,\hat q)}\leq c(\tau',\tau)\log  n 
\end{equation}
\end{enumerate}
\end{theorem}
Theorem \ref{thm:ChangeInRegime} is  a co-factor of Theorem 7 in \cite{galves:12} and its proof is presented in Appendix \ref{app:proof_changeRegime}. 
This theorem provides a criterion to choose the optimal model (and consequently, the optimal $\ell$ value) among the champion trees. That is to say, the model in $\mathcal{C}_n$ at which the change of regime occurs. This is the scope of the second step of the SMC. 

\begin{enumerate}
\item[] \textbf{Step 2. Identify the optimal tree.} 
To select an optimal tree $\hat{\tau}_{\hat{k}} \in \mathcal{C}_n$ we use the following consequence of Theorem  \ref{thm:ChangeInRegime}. For any $\tau \succeq \tau' \succeq \tau^*$,
\begin{equation}
\lim_{n \to \infty} \frac{  \log L_{(\tau, \hat{q})}\left((X,Y)_1^n\right) - \log L_{(\tau' , \hat{q})}\left((X,Y)_1^n\right) }{n} = 0.
\end{equation}
This suggest that  $\hat{\tau}_{\hat{k}} \in \mathcal{C}_n$ should be the smallest context tree such that the rescaled difference between the conditional log-likelihood of $\hat{\tau}_{\hat{k}}$ and  $\hat{\tau}_{\hat{k}-1}$ (its successor in the order $\prec$) decreases as $n$ increases. This is done by comparing average bootstrapped conditional log-likelihood using a t-test, as follows.

\begin{itemize}
\item[a)] Fix two different sample sizes $n_1 < n_2 < n$. Obtain $B$ independent bootstrap resamples of $(X_1,Y_1), ..., (X_n,Y_n)$, of size $n_2$, say
$$
(\textbf{X}^*,\textbf{Y}^*)^{(b,n_2)} = \{(X^*_1,Y^*_1)^b,...,(X^*_{n_2},Y^*_{n_2})^b\}, \quad b = 1,...,B.
$$
Similarly, let $(\textbf{X}^*,\textbf{Y}^*)^{(b,n_1)}, b = 1,...,B$ be another set of independent bootstrap samples of size $n_1$ constructed by truncating the sequences $(\textbf{X}^*,\textbf{Y}^*)^{(b,n_2)}$ to size $n_1$. 

\item[b)] For each $\hat{\tau}_k \in \mathcal{C}_n, k = K,\ldots, 2$ and its successor $\hat{\tau}_{k-1} \in \mathcal{C}_n$ ($\hat{\tau}_k \prec \hat{\tau}_{k-1}$) compute the rescaled log-likelihood differences
\begin{equation}\label{eq:stat_boot_ttest}
D^{k}_b(n_j) =
\frac{\log L_{(\hat{\tau}_k, \hat{q}_k)}\left((X^*,Y^*)^{(b,n_j)}\right) - \log L_{(\hat{\tau}_{k-1}, \hat{q}_{k-1})}\left((X^*,Y^*)^{(b,n_j)}\right)}{n_j^{0.9}},
\end{equation}
for $j = 1,2$ and $b=1,...,B$.
		
Apply a one-side t-test to compare the mean of the samples \\ $\{D^{(\hat{\tau}_k,\hat{\tau}_{k-1})}_b(n_1),  b = 1,...,B\}$ and $\{D^{(\hat{\tau}_k,\hat{\tau}_{k-1})}_b(n_2), b=1,...,B\}$.

\item[c)] Select as optimal tree $\hat{\tau}_{\hat{k}}$ the smallest champion tree such that the test rejects the equality of the means in favor of the alternative hypothesis  $E(D^{(\hat{\tau}_k,\hat{\tau}_{k-1})}(n_1)) < E(D^{(\hat{\tau}_k,\hat{\tau}_{k-1})}(n_2))$.
\end{itemize} 

\end{enumerate} 

Step 2 involves the computation of bootstrap resamples of a sequence of random objects driven by context tree models. The toolbox implements different bootstrap strategies for that. Before introducing them, we describe the bootstrap schemes implemented to resample a context tree model $(X_1,...,X_n)$.   

\begin{figure}[htb]
	\centering
	\includegraphics[width = 0.9\linewidth]{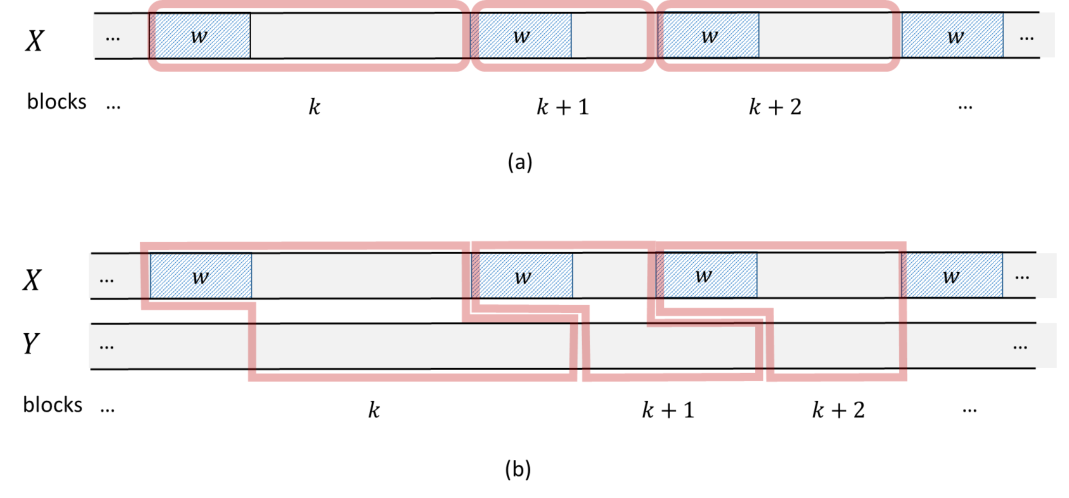}
	\caption{Illustration of a sequence split in blocks using a renewal context $w$ for (a) contex tree models and (b) sequence of random objects driven by context tree models}
	\label{fig:bootBlocks}
\end{figure}

\begin{itemize}
\item (Parametric bootstrap) The bootstrap samples are obtained by drawing from a parameterized distribution \citep{Buhlmann2002} in the following way:
\begin{enumerate}
\item [(a)]Choose a hyperparameter value $l \geq 0$ and estimate the model $(\hat{\tau}, \hat{p})$ using the data $(X_1,...,X_n)$.
\item [(b)]Generate the bootstrap samples by simulating from the approximated distribution $\hat{F} = \hat{F}_{(\hat{\tau}, \hat{p})}$,
$$
(X_1^*,...X_n^*) \sim \hat{F}_{(\hat{\tau}, \hat{p})}.
$$ 
\end{enumerate}

\item (Block bootstrap) Split the sample $(X_1,...,X_n)$ into non-overlapping blocks (see Figure \ref{fig:bootBlocks}a). These blocks are build by using a \textit{renewal context} of $X_1^n$ to split the sequence. A renewal context is a string from which the next symbols can be generated without knowing further information from the past.  
A resample is obtained by repeatedly sampling uniformly a block from the set of blocks  and concatenating them. In the toolbox, the user can specify the renewal context, or it can be computed from the estimated model $(\hat{\tau}, \hat{p})$.        
\end{itemize}

A bootstrap resampling $(X^*,Y^*)_1^n$ of the bivariate chain $(X,Y)_1^n$ can be obtained with two different procedures:
\begin{itemize}
	\item (Parametric bootstrap) 
	\begin{itemize}
		\item [(a)] Choose a hyperparameter value $\ell\geq 0$ and estimate $(\hat{\tau}, \hat{q})$ from $(X,Y)_1^n$.
		\item [(b)] Obtain a bootstrap resampling $(X^*)_1^n$ of the sequence $X_1^n$ using one of the bootstrap strategies for context tree models described above. The user can also choose not to resample the sequence $X_1^n$.
		\item [(c)] Generate a sequence $(Y^*)_1^n$ using the distribution $\hat{q}(\cdot|c_{\hat{\tau}}((X^*)_1^n))$. 
	\end{itemize}
	   
	\item (Block bootstrap) Split the sample $(X_1,Y_1),...,(Y_n,X_n)$ into non-overlapping blocks using a renewal context (see Figure \ref{fig:bootBlocks}b). In this case, a renewal context is a string from $X_1^n$ such that from it is possible to generate both the next symbols of sequence $(X_n)_n$ and its associates responses $(Y_n)_n$, without further information from the past. The bootstrap samples can be obtain by repeatedly  sampling uniformly from the set of blocks and concatenating them.    
	
\end{itemize}

\begin{remark}
	\cite{Buhlmann2000} proposed a procedure based on Risk functions for tuning the algorithm Context. The SeqROCTM toolbox implements this tuning procedure for the particular case of the Final prediction error risk. This procedure is available in the toolbox for tuning the model selection procedures for context tree models and the model selection procedures for sequence of random objects driven by context tree models (for the categorical case).
\end{remark}

\section{Software Architecture}
\label{sec:soft_framework}

\begin{figure}[htb]
	\centering
	\includegraphics[width = .9\linewidth]{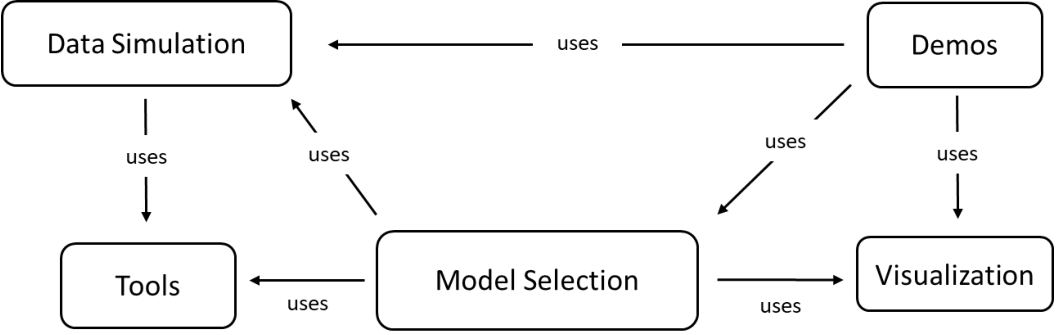}
	\caption{The software architecture of the Matlab SeqROCTM toolbox.}
	\label{fig:architecture}
\end{figure}

The SeqROCTM toolbox have been designed following a modular structure. The toolbox consists of functions written in Matlab that can be grouped in four modules regarding their functionalities (see Figure \ref{fig:architecture}). This architecture makes the software easy to update, either by adding new functionalities or by improving the existing ones. 

The module Data Simulation includes routines to simulate sequences of inputs (i.e., context tree models), to simulate the response sequence of a sequence of random objects driven by context tree models and to simulate the bivariate sequence. The module Visualization implements the algorithm described in \cite{Mill} to graphically show a tree structure. This module contains also a routine to print the context tree in the console. The Tools module implements several functions that can assist the researcher during the experimental design and data analysis. Some of those functions are also invoked by functions in other modules. Some demos illustrating how to use the toolbox are included in the Demo module. 

\begin{figure}[htb]
	\centering
	\includegraphics[width = \linewidth]{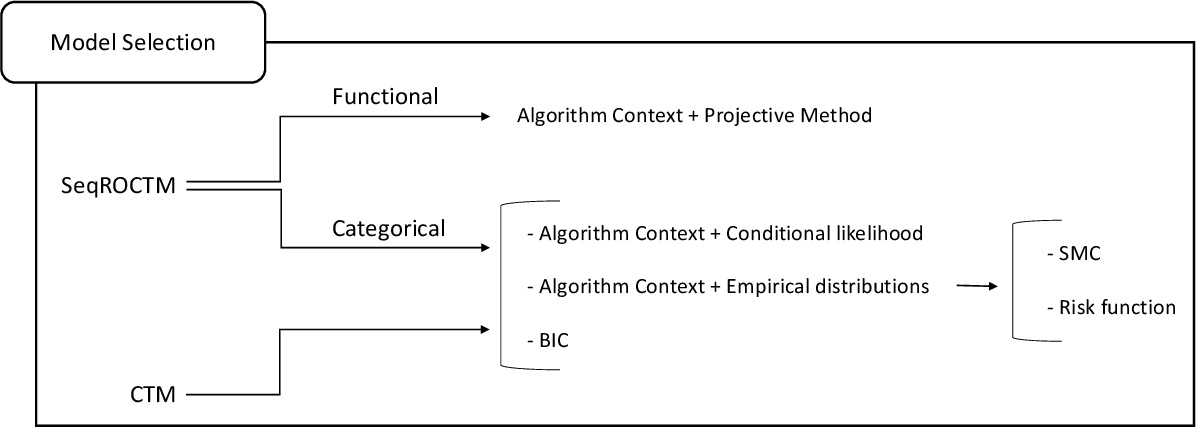}
	\caption{The different algorithms included in the Model Selection module of the SeqROCTM toolbox.}
	\label{fig:modelselectionmodule}
\end{figure}

The main functions of the toolbox are in the Model Selection module. This module contains all the model selection procedures and tuning algorithms introduced in Section \ref{sec:modelselection}. Figure \ref{fig:modelselectionmodule} presents a close-up of this module.

The novelty implemented in this toolbox it is illustrated in Figure \ref{fig:modelselectionmodule} by the branch growing from the SeqROCTM node, that is, the mathematical framework to make inference in the class of sequences of random objects driven by context tree models. Nevertheless, due to the close relation to model selection in context tree models, we end up including in the toolbox several existing model selection algorithms for context tree models. 

Up to our knowledge, there is only an R-package implementing model selection in context tree models introduced by \cite{Machler2004}. This package implements the algorithm Context and a tuning procedure for the algorithm Context based on Risk functions. Our toolbox also works as an alternative tool for this purpose. 

All the implementations are self-contained. The only external function the toolbox uses is the function permn \citep{permn}. 


\section{Illustrative example: The Goalkeeper game}
\label{sec:example1}

This section presents the major functionalities of the SeqROCTM toolbox through an illustrative example. 

The Goalkeeper game is a video game developed by the NeuroMat team (\url{https://game.numec.prp.usp.br/}) as a tool to investigate the conjecture that the brain does statistical model selection \citep{Castro2006}. During the game, the kicker can throw the ball in three directions: \textit{left}, \textit{center} or \textit{right}. The agent, playing the role of the Goalkeeper in a soccer penalty shootout, has to defend as much penalties as possible by predicting, at each step, in which direction the kicker will throw the ball. The kicker's choices are randomly generated according to a context tree model. The Goalkeeper game has been used in the experimental protocol of some neurobiological experiments \citep{stern2020}.

Here, for simplicity, instead of collecting some data using the game, we will simulate different participant strategies to generate the responses. We show how the toolbox can be used to assess the participant strategy from the data. The aim is that the estimated strategy matches the one used to generate the responses matches.   

To generate the sequence of kick directions, we define a context tree model that generates sequences according to the following rule: after a shot to the \textit{left} the kicker always send the ball to the \textit{center}, after a shot to the \textit{right} he always send the ball to the \textit{left}, but after a shot to the \textit{center}, if one step back he sent the ball to the \textit{center}, then he send the ball to the \textit{left}. Otherwise, if one step back he sent the ball to the \textit{left}, then he send the ball to the \textit{right} with probability 0.8 and to the \textit{center} with probability 0.2. 

Formally, the directions \textit{left}, \textit{center} and \textit{right} are represented by the symbols 0, 1 and 2, respectively. These symbols will conform the alphabet. When using the toolbox, an alphabet is always represented by a vector of consecutive positive integers from zero to the number of elements minus one. A context tree is defined through a cell array of vectors, each vector representing a context (i.e., a leaf of the tree). The distributions associated to the contexts are specified by a matrix with the number of rows equals the number of contexts and the number of columns equals the number of symbols in the alphabet. In this way, the $k$-th row contains the distribution associated to the $k$-th context in the cell array defining the context tree. The following source code defines these variables according to the example. 

\begin{lstlisting}
	% alphabet of three symbols
	A = [0,1,2];
	
	% context tree containing four contexts 0, 2, 01, 11
	tau = {0, 2, [0,1], [1,1]};
	
	% distributions associated to each contexts (4x3 matrix)
	% e.g., first row indicates the distribution of context 0, that is p(0|0)=0, p(1|0)=1, p(2|0)=0
	p = [0, 1, 0 ; 1, 0, 0; 0, 0.2, 0.8; 1, 0, 0 ];
	
	% visualize the context tree
	draw_contexttree(tau, A);
\end{lstlisting}

To generate a sequence of stimuli according to a given context tree model we use the function \texttt{generatesampleCTM}, which receives as inputs a context tree, the probability distributions associated to the contexts, an alphabet and the length of the sequence we want to generate. The code below generates a sequence of inputs of length 300, using the variables already defined.  

\begin{lstlisting}
	% length of the sequence
	seq_length = 300;
	% sequence of stimuli (context tree model) in a row vector
	X = generatesampleCTM(tau, p, A, seq_length);
\end{lstlisting}

The variable \texttt{X} contains the sequence of the kicker choices. We will define three different strategies for the goalkeeper and generate a response sequence for each of them, say \texttt{Y1}, \texttt{Y2} and \texttt{Y3}. When applying the model selection procedures to the data \texttt{(X1,Y1)}, \texttt{(X2,Y2)} and \texttt{(X3,Y3)}, the desired result is to recover the strategy used to simulate the goalkeeper responses on each case. 

The three strategies used to simulate the goalkeeper responses are the following:
\begin{itemize}
	\item Strategy 1. Every time the goalkeeper see the shot of the kicker to the left, he will defend the next shot to the center. After a shot to the center, the goalkeeper will defend to the right. And after a shot to the right, the goalkeeper will defend to the left. Using the variables already defined, this strategy can be translated as follows: If $X_n = 0$, then $Y_{n+1} = 1$. If $X_n = 1$, then $Y_{n+1} = 2$. If $X_n = 2$, then $Y_{n+1} = 0$.
	\item Strategy 2. The goalkeeper learns the relevant pasts (i.e., the contexts) of the sequence $X$ and, at each step, he identifies the context associated to the current past and chooses the direction with grater probability of being generated after that context. This is the strategy that maximizes the probability of matches. This means that whenever the goalkeeper see a shot to the center preceded by a shot to a left, he will defend the next shot to the right. On the contrary, if the shot to the center is preceded by another shot to a center, the goalkeeper will defend the next shot to the left. Besides, if the kicker shot the ball to the left, the goalkeeper will defend the next shot to the center and if the kicker shot the ball to the right, the goalkeeper will defend the next ball to the left. This means that if $X_n = 0$, then $Y_{n+1} = 1$. If $X_n = 1$ and $X_{n-1} = 0$, then $Y_{n+1} = 2$. If $X_n = 1$ and $X_{n-1} = 1$, then $Y_{n+1} = 0$. If $X_n = 2$, then $Y_{n+1} = 0$.     
	\item Strategy 3. The goalkeeper pays no attention to the temporal dependences encoded in the kicker strategy, at each step, randomly chooses in an independent and uniform way \textit{left}, \textit{center} or \textit{right}.  
\end{itemize}

The source code below defines the context tree and the distributions used to simulate the goalkeeper responses according to the described strategies. To generate the response sequence, the function \texttt{generatesampleYSeqROCTM} is used. 

For the categorical case, which is the case of the current example, the observed sequence of responses must be stored in a row vector. For the functional case, the response sequence must be specified by a matrix containing on each column a chunk of function (this will be exemplified later, in the illustrative example presented in Section 4).

\begin{lstlisting}
% Strategy 1
ctx1 = {0, 1, 2};
q1 = [0 1 0; 0 0 1; 1 0 0];
[X1, Y1] = generatesampleYSeqROCTM(X, ctx1, q1, A);

% Strategy 2
ctx2 =  {0, 2, [0,1], [1,1]};
q2 = [0, 1, 0 ; 1, 0, 0; 0, 0, 1; 1, 0, 0 ]; 
[X2, Y2] = generatesampleYSeqROCTM(X, ctx2, q2, A);

% strategy 3
ctx3 =  {};
q3 = [1/3 ; 1/3; 1/3 ]; 
[X3, Y3] = generatesampleYSeqROCTM(X, ctx3, q3, A);
\end{lstlisting}

Now that we have some data, i.e., \texttt{(X1,Y1)}, \texttt{(X2,Y2)} and \texttt{(X3,Y3)}, we will exemplified how the functions responsible for model selection can be used. For the current example, we will use the function \texttt{tune\_SeqROCTM}, which receives as mandatory inputs the data and the alphabet. There exists a lot of optional name-value pairs arguments, which could be specified also as input of the function. The following source code shows how to invoke the \texttt{tune\_SeqROCTM} function specifying a different estimation method for each data and SMC as tuning procedure for all the cases.  

\begin{lstlisting}
% some parameters value
c_min = 0;
c_max = 1000;     % high enough, such as to obtain the empty tree
max_height = 6;
alpha = 0.05;

% tune the SeqROCTM model for each strategy
[~,~, r1] = tune_SeqROCTM(X1, Y1, A, 'TuningMethod', 'smc', 						 ...
									 				  'EstimationMethod', 'context_cL', ...
									 				  'MaxTreeHeight', max_height, 	 		 ...
									 				  'ParameterLowerBound', c_min,  		 ...
								     				  'ParameterUpperBound', c_max,  		 ...
									 				  'Alpha', alpha);

[~,~, r2] = tune_SeqROCTM(X2, Y2, A, 'TuningMethod', 'smc',             		   ...
													  'MaxTreeHeight', max_height,       		...
													  'EstimationMethod', 'context_empD',	...
													  'ParameterLowerBound', c_min,      		...
													  'ParameterUpperBound', c_max,      		...
													  'Alpha', alpha);

[~,~, r3] = tune_SeqROCTM(X3, Y3, A, 'TuningMethod', 'smc',             	 ...
													  'MaxTreeHeight', max_height,       ...
													  'EstimationMethod', 'bic',         ...
													  'ParameterLowerBound', c_min,      ...
													  'ParameterUpperBound', c_max,      ...
													  'Alpha', alpha,                    ...
													  'BootNsamples', 200,               ...
													  'BootStrategy', 'blocks');
												      
% show the results of the estimation procedures
figure
for i = 1 : 3
subplot(2,3,i)
% get the structure of the corresponding model
eval(['r = r' num2str(i) ';']); 
% get the values from the structure r
nleaves = cellfun(@(x) size(x,2), r.champions);
ML = r.fvalues;
idtree = r.idxOptTree;
cutoff = r.prmvalues;
% draw the curve
plot(nleaves, ML, '*--b')
hold on; plot(nleaves(idtree), ML(idtree), 'ro');
text(nleaves(idtree)+0.5, ML(idtree), ['\leftarrow C = ' num2str(cutoff(idtree))], 'FontSize', 8);
ylabel('$$\log(L_{(\tau, \hat{q})}(Y_1^n|X_1^n))$$', 'interpreter', 'latex');
xlabel('$$|\tau|$$', 'interpreter', 'latex');
% draw the choosen context trees
subplot(2,3,3+i)
draw_contexttree(r.champions{idtree}, A, [1 0 0], 3);
end

% Calling the model selection procedure without tuning (using the default
% value of the hyperparameter)
[tau1, q1] = estimate_discreteSeqROCTM(X1, Y1, A, 'MaxTreeHeight', max_height, 'EstimationMethod', 'context_empD');
[tau2, q2] = estimate_discreteSeqROCTM(X2, Y2, A, 'MaxTreeHeight', max_height, 'EstimationMethod', 'context_cL');
[tau3, q3] = estimate_discreteSeqROCTM(X3, Y3, A, 'MaxTreeHeight', max_height, 'EstimationMethod', 'bic');

% show the results in the console
print_tree(tau1);
print_tree(tau2);
print_tree(tau3);

\end{lstlisting}

\begin{figure}[htb]
	\centering
	\includegraphics[width = \linewidth]{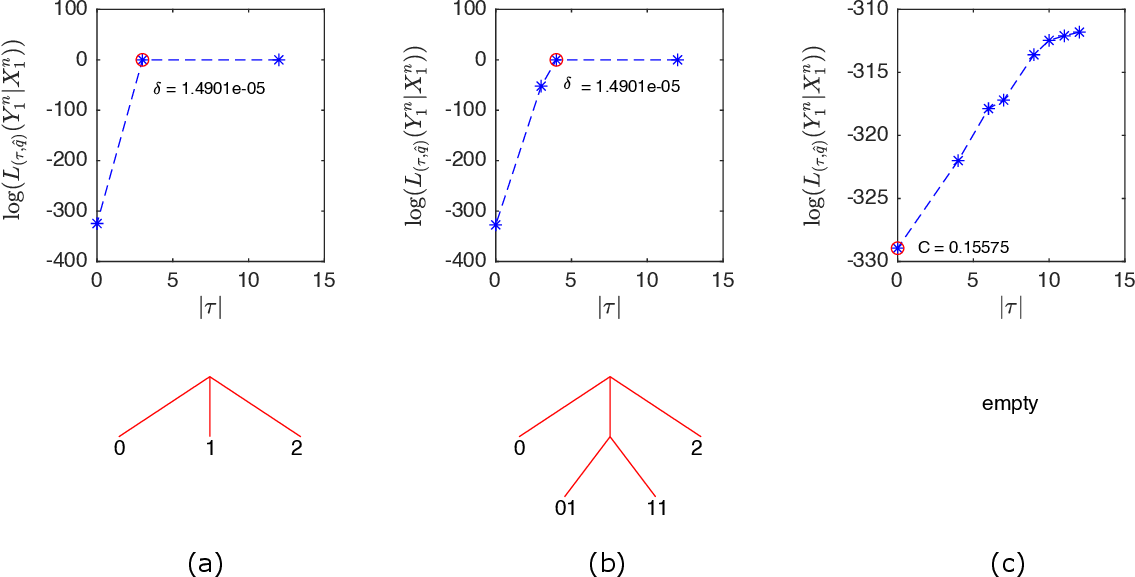}
	\caption{Result of the tunning procedure for a) strategy 1 b) strategy 2 and c) strategy 3. The first row shows plots of the logarithm of the conditional likelihood vs. the number of contexts for each model in the set of champion trees. The red circle indicates the optimal model chosen using SMC. The second row shows the context tree corresponding to the optimal model.}
	\label{fig:tuning}
\end{figure}

Figure \ref{fig:tuning} shows the results obtained for each simulated strategy (this Figure is also generated by the code above). For each goalkeeper strategy it is shown the conditional log-likelihood of each champion tree as a function of its size. The optimal model chosen using SMC is marked with a red circle and the corresponding context tree is graphically shown below. The optimal value of the hyperparameters ($\delta$ in the first two cases and $c$ in the third one) is also shown in the plot. For all the strategies, the model estimated from the data using the tunning procedure matches the context tree used to simulate the goalkeeper responses.     

The source code also shows how to invoke the function \texttt{estimate\_discreteSeqROCTM}, which is responsible for model selection without any tuning procedure. The user can specify as input a value for the hyperparameter (as it was done for strategy 3). If no value for the hyperparameter is given as input, a default value is used. 

\section{Illustrative example: Retrieving the structure of probabilistic sequences from EEG data}
\label{sec:example2}

Humans are great at learning statistical regularities from sequences of stimuli. Having learned patterns from the inflow of sensory information, one can predict the upcoming stimuli to improve perception and decision making \citep{summerfield_expectation_2014,de_lange_how_2018}. Understanding the capacity of the brain to learn statistical regularity from temporal sequences has been the focus of several researches in neuroscience. This was the focus of the recently published study by \cite{Hernandez:21} from which we extracted the current illustrative example .

Consider a sequence of auditory stimuli generated by a context tree model. This sequence is presented to a volunteer while electroencephalographic (EEG) signals are recorded from his scalp. In this framework, the conjecture that the brain identifies statistical regularities from sequences of stimuli can be rephrased by claiming that the brain identifies the context tree used to generate the sequence of auditory stimuli. If this is the case, a signature of the context tree should be encoded in the brain activity. The question is whether this signature can be identified in the EEG data recorded during the experiment. 

The auditory units used as stimulus are either {\it strong beats}, {\it weak beats} or {\it silent units}, represented by symbols $2, 1$ and $0$, respectively. The statistical regularity encoded in the sequences of stimuli can be informally described as follows. Start with the deterministic sequence
$$ 2 \ 1 \ 1\ 2 \ 1 \ 1 \ 2 \ 1 \ 1 \ 2 \ 1 \ 1 \ 2 \ldots , $$ then replace each weak beat (symbol $1$) by a silent unit (symbol $0$) with a small probability, say 0.2, in an independent way. An example of a sequence produced in this way would be
$$
2 \ 1 \ 1\ 2 \ 0 \ 1 \ 2 \ 1 \ 1 \ 2 \ 0 \ 0 \ 2\ldots .
$$
This stochastic sequence constitutes a sample of a context tree model compatible with the context tree and the family of transition probabilities shown in Figure \ref{fig:ternary}.
\begin{figure}[htb]
	\begin{minipage}[c]{0.4\textwidth}
		\centering
		\begin{tikzpicture}[thick,scale=1]
		\tikzstyle{invisible} = [rectangle, node distance=0.5cm]
		\tikzstyle{level 1}=[level distance=1.3cm, sibling distance=2.2cm]
		\tikzstyle{level 2}=[level distance=1.2cm, sibling distance=0.8cm]
		\coordinate
		child{{}
			child {[fill] circle (2.5pt) node (00) {} }
			child {[fill] circle (2.5pt) node (10) {} }
			child {[fill] circle (2.5pt) node (20) {} }		        	
		}
		child{ {} 
			child {[fill] circle (2.5pt) node (01) {} }
			child {[fill] circle (2.5pt) node (11) {} }
			child {[fill] circle (2.5pt) node(21) {} }        
		}
		child{[fill] circle (2.5pt) node(2){} }
		;
		\node [ invisible, below of=00 ]{00};
		\node [ invisible, below of=10 ]{10};
		\node [ invisible, below of=20 ]{20};
		\node [ invisible, below of=01 ]{01};
		\node [ invisible, below of=11 ]{11};
		\node [ invisible, below of=21 ]{21};
		\node [ invisible, below of=2 ]{2};
		\end{tikzpicture}
	\end{minipage} \qquad
	\begin{minipage}[c]{0.4\textwidth}
	\scalebox{0.8}{
		\begin{tabular}{c c c c}
			\toprule
			\textbf{context} $\mathbf{w}$ & 
			$\mathbf{p(0 | w)}$ & 
			$\mathbf{p(1 | w)}$ & 
			$\mathbf{p(2 | w)}$\\
			\toprule
			2   &  0.2 & 0.8 & 0\\
			21 &  0.2 & 0.8 & 0\\
			20 &  0.2 & 0.8 & 0 \\
			11 &  0 & 0 & 1\\
			10 &  0 & 0 & 1 \\
			01 & 0 & 0 & 1\\
			00 &  0 & 0 & 1\\
			\bottomrule
	\end{tabular}}
\end{minipage}
	\caption{Graphical representation of the context tree and the transition probabilities associated to the contexts.}
     \label{fig:ternary}
\end{figure}


To obtain a context tree from the EEG data the statistical model selection procedure for sequences of random objects driven by context tree models introduced for the functional case is employed. This procedure is applied separately to each participant data. Participants are not exposed exactly to the same sequence of stimuli, but different realizations of the same context tree model. 


This illustrative example presents a tiny part of a wider experimental protocol introduced in \cite{Hernandez:21}. The experimental protocol involves 19 participants, two different context tree models to generate the sequences of stimuli and 18 electrodes in which the EEG data is recorded. To show how the SeqROCTM was used here, we will consider only the EEG signals recorded in one electrode for 3 participants.  

We start by exemplifying how to generate sequences of stimuli of length 700 using the context tree model of Figure \ref{fig:ternary}. 

\begin{lstlisting}
% number of volunteers
n_volunteers = 3;

% alphabet and context tree model used to generate the sequence of stimuli
A = [0,1,2];
tau = {[0,0], [1,0], [2,0], [0,1], [1 1], [2,1], 2};
p = [0, 0, 1 ; 0, 0, 1; 0.2, 0.8, 0; 0, 0, 1; 0, 0, 1; 0.2, 0.8, 0; 0.2, 0.8, 0];

% length of the sequences of stimuli
seq_length = 700;

% Sequences of stimuli
% matrix X of 3x700 containing on each row a sequence of stimuli
Xdata = zeros(3,700);
for v = 1 : n_volunteers
	Xdata(v,:) = generatesampleCTM(tau, p, A, seq_length);
end
\end{lstlisting}

In the following, we load the EEG data recorded from a frontal electrode (FP1) for 3 participants as well as the sequences of stimuli the participants were exposed to. This EEG data is already pre-processed and segmented. The pre-processing details are omitted because are out of the scope of this article. 

\begin{lstlisting}

% load sequence of stimuli and EEG data for each volunteer
names_volunteer = {'V02', 'V09', 'V19'};

X = [];
Y = cell(1,3);

for v = 1 : n_volunteers

	% load stimuli data 
	vname_i = [names_volunteer{v} '_stimuli'];
	x = load(vname_i);
	x = x.(vname_i);
	X = [X; x];

	% load response data
	vname_r = [names_volunteer{v} '_response'];
	y = load(vname_r);
	y = y.(vname_r);
	Y{v} = y;
end

% visualize some symbols of the stimuli sequence and its corresponding EEG
% chunks for volunteer V02
figure;
id_cols = 760:768;
for i = 1 : 9
	% plot the stimuli
	ax = subplot(2, 9, i);
	text(0.5, 0.5, num2str(X(1, id_cols(i))), 'FontSize', 20);
	set( ax, 'visible', 'off')

	% plot the EEG chunk
	ax = subplot(2, 9, 9+i);
	plot(Y{1}(:, id_cols(i)));
	set( ax, 'visible', 'off')
	xlim([0 115])
end

\end{lstlisting}

\begin{figure}[htb]
	\centering
	\includegraphics[width = 0.9\linewidth]{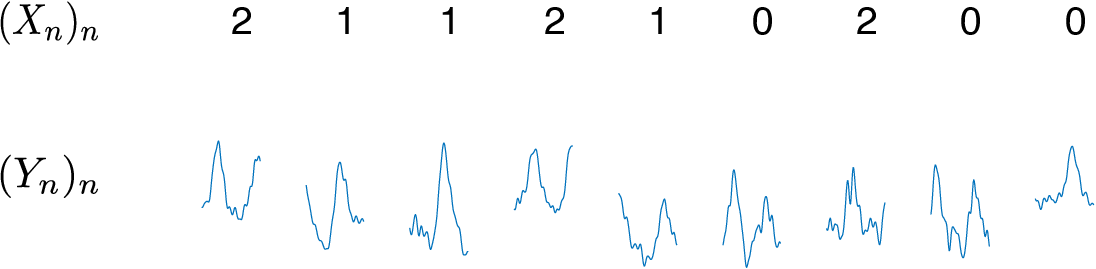}
	\caption{Symbols in the sequence of stimuli and their corresponding EEG chunks for participant V02.}
	\label{fig:seq}
\end{figure}

Figure \ref{fig:seq} shows some elements of the sequence of stimuli and the corresponding EEG chunks for the first volunteer.

We are now ready to invoke the model selection functions. The values of the parameters required by this function are specified in the source code.   

\begin{lstlisting}
% model selection algorithm on the data of each volunteer
nBM = 1000;
Alpha = 0.05;
Beta = 0.05;

rng(1); tree_v02 = estimate_functionalSeqRoCTM(X(1,:), Y{1}, A, 3, nBM, Alpha, Beta, 0);
rng(1); tree_v09 = estimate_functionalSeqRoCTM(X(2,:), Y{2}, A, 3, nBM, Alpha, Beta, 0);
rng(1); tree_v19 = estimate_functionalSeqRoCTM(X(3,:), Y{3}, A, 3, nBM, Alpha, Beta, 0);

% draw the results
figure
subplot(1,3,1)
draw_contexttree(tree_v02, A, [1 0 0], 3);
subplot(1,3,2)
draw_contexttree(tree_v09, A, [0 1 0], 3);
subplot(1,3,3)
draw_contexttree(tree_v19, A, [0 0 1], 3);
\end{lstlisting}

\begin{figure}[htb]
	\centering
	\includegraphics[width = 0.8\linewidth]{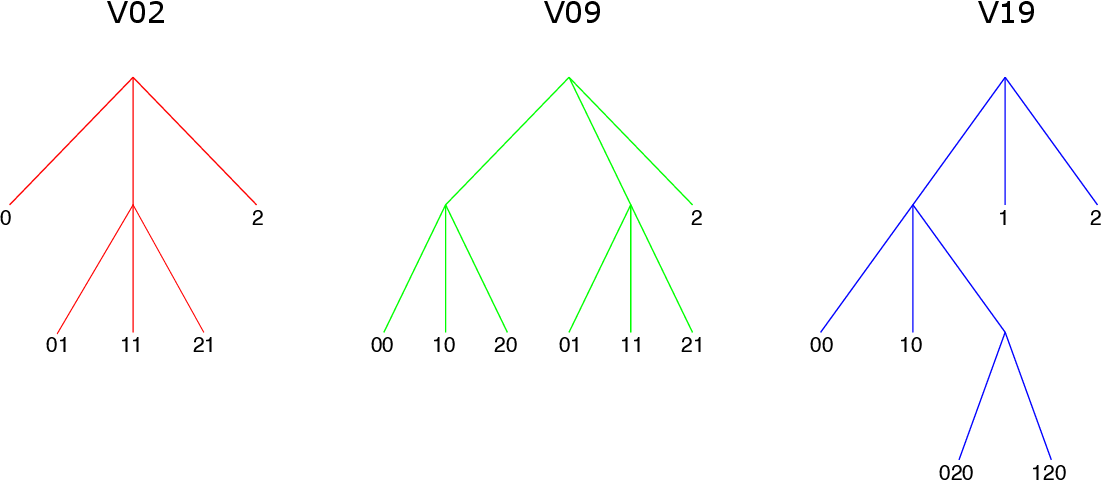}
	\caption{Context tree retrieved for each participant.}
	\label{fig:context_trees_volunteers}
\end{figure}

Figure \ref{fig:context_trees_volunteers} shows the context tree retrieved from the EEG data of each participant. 

It can be seen that for one of the participants the retrieved context tree is the same that the one generating the sequence of stimuli. For the other two participants, the recovered tree differs from the stimuli tree by one branch. More details about this experiment can be found in the Discussion section of \cite{Hernandez:21}.

\section{Conclusions}
\label{sec:conclusions}

This paper introduces the Matlab SeqROCTM toolbox aimed to implement model selection procedures in a new class of stochastic process, namely sequences of random objects driven by context tree models. This is a new mathematical framework that finds nice applications in several scientific fields such as neuroscience, linguistic, genetics. The toolbox also implements different procedures for model selection of context tree models.
We have described the main functionalities of the toolbox and we give examples of how to use it.    

Further extensions are planned in order to solve some limitations of the current version, e.g., greater flexibility when defining the alphabet. 
For simplicity, in the categorical case we restrict the finite set in which $Y_n$ takes values to the set $A$, where the associated context tree model takes value. 

\section{Acknowledge}

The authors thanks professor Antonio Galves from University of S\~{a}o Paulo for the discussions and suggestions that contribute to this article.    

\appendix
\section{Proof of Proposition \ref{prop:binomial}}
\label{app: proof-binomial}
\begin{proof}
Given a string $u\in A^*$, for each $a,b\in  \tau_u$, we set
$$
Z^{a,b}_{i,n}=D^{W_i}_n\big((Y_1^{(a)},\ldots, Y_{N_n(a)}^{(a)}),(Y_1^{(b)},\ldots, Y_{N_n(b)}^{(b)})\big).
$$
Assume that the null assumption $H^{(u)}_0$ is true.  
Then the asymptotic properties of the Kolmogorov-Smirnov statistics implies, for each $a,b\in  \tau_u$,  with $a\neq b$ and $1\leq i\leq N$,  that $Z^{a,b}_{i,n}$ converges in distribution to $K=\sup_{t\in[0,1]} |B(t) |$ as $n\to\infty$, where $B=(B(t):t\in [0,1])$ is a  Brownian bridge. 

Let $c_\alpha=\sqrt{-1/2 \ln(\alpha/2)}$ be the $\alpha-$percentile of the Kolmogorov distribution. When $|\tau_u|= 2$, let say $\tau_u=\{a,b\}$,  it holds that $P(\Delta^{W_i}_n(u)>c_\alpha )=P(Z^{a,b}_{i,n}>c_\alpha)= \alpha$ as $n\to \infty$. 
If  $|\tau_u|> 2$, define  $M= {|\tau_u|\choose 2}$ and  in this case $P(\Delta^{W_i}_n(u)>c_{\alpha/M})=P(\cup_{a,b\in \tau_u\in} Z^{a,b}_{i,n}>c_{\alpha/M})=\bar{\alpha}\leq \alpha$ as $n\to \infty$. Therefore, taking $\delta_\alpha(u) = c_{\alpha/M}$, it holds that 
\begin{equation}
P( \Delta^{W_i}_n(u)>\delta_\alpha(u) )\leq \alpha, \ \ \mbox{as  } n\to \infty .
\end{equation}

In what follows,  for each $1\leq i\leq N$, we define 
$$
Z_{i,n}=1\{ \Delta^{W_i}_n(u) > \delta_\alpha(u) \}.
$$
We will show that for any $a_1,\ldots,a_N\in\{0,1\}$,
\begin{equation}
\label{converge_distr}
\lim_{n\to \infty}P(Z_{i,n}=a_i,\ldots, Z_{i,N}=a_N)=\bar\alpha^{\sum_{i=1}^N a_i}(1-\bar\alpha)^{(N-\sum_{i=1}^N a_i)}
\end{equation}
where $\bar\alpha$ denotes either $\alpha$, if $|\tau_u|=2$, or $P(\cup_{a,b\in \tau_u\in} Z^{a,b}_{i,n}>c_{\alpha/M})$, if $|\tau_u|>2$.

Denote $\mathcal{G}=\sigma(Y^{(as)}_k,k\geq 1,a\in A)$ and notice that conditionally on $\mathcal{G}$, the random variables $Z_{1,n}, \ldots, Z_{N,n} $ are independent for all  $n\geq 1$.
By the Skorohod's representation theorem, there is a sequence of random vectors $(\tilde{Z}_{1,n},\ldots, \tilde{Z}_{N,n})_{n\geq 1}$ and a sequence of random elements $(\tilde{Y}^{(as)}_k)_{k\geq 1,a\in A}$ taking values in $L^{2}([0,T])$, both sequences defined in the same probability space $(\tilde{\Omega},\tilde{\mathcal{F}},\tilde{P})$  such that 
\begin{enumerate}
\item  for each $n$, $(\tilde{Z}_{1,n},\ldots, \tilde{Z}_{N,n})$ has the same distribution as $(Z_{1,n},\ldots, Z_{N,n})$,
\item for each $k$ and $a\in A$, the distribution of $\tilde{Y}^{(as)}_k$ is the same as the distribution of $Y^{(as)}_k.$ 
\item if $\tilde{\mathcal{G}}=\sigma(\tilde{Y}^{(as)}_k, k\geq 1, a\in A)$, then $\tilde{Z}_{1,n},\ldots, \tilde{Z}_{N,n}$ are conditionally independent given $\tilde{\mathcal{G}}$,
\item for each $1\leq i\leq N$, $\tilde{Z}_{i,n}\to K$ almost surely with respect to $\tilde{P}$ as $n\to\infty$. 
\end{enumerate}
Item 4 and the Dominate convergence theorem for conditional expectation imply that $\tilde{P}$-a.s as $n\to\infty$, for each $1\leq i\leq N$ and $a_i\in\{0,1\}$,
$$
\tilde{P}(\tilde{Z}_{i,n}=a_i|\tilde{\mathcal{G}})\to \bar \alpha^{a_i}(1-\bar \alpha)^{(1-a_i)}.
$$  
Therefore, by Item 3 and the Dominate convergence theorem, we have that for any $a_1,\ldots,a_N\in\{0,1\}$, as $n\to \infty$,
\begin{eqnarray*}
\tilde{P}(\cap_{i=1}^{N}\tilde{Z}_{i,n}=a_i)&=&\tilde{E}\left[\prod_{i=1}^N\tilde{P}(\tilde{Z}_{i,n}=a_i |\tilde{\mathcal{G}})\right]\to \bar \alpha^{\sum_{i=1}^N a_i}(1-\bar \alpha)^{(N-\sum_{i=1}^N a_i)} 
\end{eqnarray*}
The limit in \eqref{converge_distr} now follows from Item 1. 

Finally, Proposition \ref{prop:binomial} follows from the fact that,  given two random variables $\bar\eta$ and $\eta$ with Binomial distribution of parameters $(N,\bar \alpha)$ and $(N,\alpha)$, respectively. If $\bar \alpha\leq \alpha,$ then $P(\bar{\eta}>k)\leq P(\eta>k)$.

\end{proof}

\section{Proof of Proposition \ref{thm:tau_in_Cn}}\label{app:proof_orderedChTrees}
\begin{proof}
For the BIC case the proof is a direct adaptation of Theorem 6 in \cite{galves:12} using Remark \ref{rmk:BIC_Xn} and the conditional log-likelihoods $L_{(\hat \tau, \hat q)}\left((X,Y)_1^n\right)$ instead of $L_{(\hat \tau, \hat p)}(X^n_1)$.

Algorithm Context is strongly consistent either when using conditional log-likelihood \citep{Rissanen:83} or offspring empirical distributions \citep{Galves-Leonardi2008} (see Remarks \ref{rmk:consistency_AC} and \ref{rm:consis_mod_AC}) and since the set of champion trees is countable we get the first part of the proposition.
Therefore, it remains to show only the ordering of the champion trees with respect to $\succ$ for both cases.

We shall do the proof for the algorithm Context with conditional log-likelihood. The proof for algorithm Context with offspring distributions its the same (replacing $\tilde{\Delta_n}$ by ${\Delta_n}$).
Given $0<\delta_1<\delta_2$, denote by $\tau^i=\hat \tau^{\delta_i}_{C,n}$ for $i=1,2$. If $\tau^1=\emptyset$ then for any $w$ with $N^X_n(w)\geq 1$ and $l(w)\leq L$ it holds that $\Delta(w)<\delta_1<\delta_2$ and therefore $\tau^2=\emptyset$.

On the other hand, if $\tau^1\neq \emptyset$,  then for any $w\in \tau^1$ it is enough to show that either $w\in \tau^2$ or there exists $w' \in \tau^2$ such that $w'\prec w$.
By  \eqref{eq:tree_ctxt}, once $w\in \tau^1$,  for any $s\in A^*$ such that $N_n^X(s)\geq 1$, $l(s)\leq L$ and $s\succeq w$, we have $\Delta(s)<\delta_1<\delta_2$.  Therefore, no string $s\succ w$ belongs to $\tau^2$, which implies that either $w\in \tau^2$ or there exists $w'\in \tau^2$ such that $w'\preceq w$.
\end{proof}

\section{Proof of Theorem \ref{thm:ChangeInRegime}}\label{app:proof_changeRegime}
The proof is a straight adaptation of Theorem 7 in \cite{galves:12} for the case of conditional log-likelihoods. 
\begin{proof}
To show the (1) consider any $\tau \in \mathcal{C}_n$ satisfying $\tau \prec \tau* $ and notice that
\begin{eqnarray*}
\lefteqn{ \log L_{(\tau^*,q)}-\log  L_{(\tau,q)}  \qquad} \\
&& = \sum_{w^*\in \tau^*}  \sum_{a\in A} N^{XY}_n(w^*,a) \log \hat q(a| w^*)- \sum_{w\in \tau}  \sum_{a\in A} N^{XY}_n(w,a) \log \hat q(a| w)\\
 && = \sum_{w\in \tau} \sum_{\substack{ {w^*\in \tau^*}\\{w\prec w^*} } } \,  \sum_{a\in A} N^{XY}_n(w^*,a) \log  \hat q(a| w^*)) - \sum_{w\in \tau}  \sum_{a\in A} N^{XY}_n(w,a) \log \hat q(a| w).
\end{eqnarray*}
Dividing both sides by $n$, the ergodic theorem implies that $N^{XY}_n(w^*,a)/n\to q(w^*, a)$, therefore, as $n\to \infty$, the right hand side of equation above, converges to
\begin{eqnarray} \label{eq:jensen}
  \sum_{w\in \tau} \sum_{\substack{ {w^*\in \tau^*}\\{w\prec w^*} } } \,  \sum_{a\in A} {q}(w^*, a) \log   q(a| w^*) -   \sum_{w\in \tau}  \sum_{a\in A} {q}(w, a) \log   q(a| w) .
\end{eqnarray} 

Now,  Jensen's inequality implies that
\begin{equation} \label{eq:jensen2}
q(w)\sum_{\substack{ {w^*\in \tau^*}\\{w\prec w^*} } }  \frac{q(w^*)}{q(w)} \big(  q(a|w^*) \log q(a| w^*)  \big)\geq q(wa)\log q (a|w), \  a \in A,
\end{equation} 
and the equality only holds if $q(a|w)=q(a|w^*)$ for each $a\in A$ and $\tau \in w\prec w^*\in \tau^* $, which is a contradiction with the minimality of $\tau^*$. Therefore there exists at least one symbol $a\in A$ such that the strict inequality holds.  Thus, applying inequality \eqref{eq:jensen2} in the left term of \eqref{eq:jensen} we conclude that must be strict positive.

To prove (2), observe that
\begin{eqnarray*}
\lefteqn{ \log L_{(\tau',q)}-\log  L_{(\tau,q)}  \qquad} \\
&& = \sum_{w'\in \tau'}  \sum_{a\in A} N^{XY}_n(w',a) \log \hat q(a| w')- \sum_{w\in \tau}  \sum_{a\in A} N^{XY}_n(w,a) \log \hat q(a| w)\\
&& \leq  \sum_{w'\in \tau'}  \sum_{a\in A} N^{XY}_n(w',a) \log \hat q(a| w')- \sum_{w\in \tau}  \sum_{a\in A} N^{XY}_n(w,a) \log  q^*(a| w)\\
 && = \sum_{w\in \tau}  \sum_{\substack{ {w'\in \tau}\\{w\prec w'} } }  \sum_{a\in A} N^{XY}_n(w',a) \log \Big( \frac{\hat q(a| w')}{ q^*(a| w)} \Big)\\
&& = \sum_{w\in \tau}  \sum_{\substack{ {w'\in \tau}\\{w\prec w'} } } N^{X}_n(w') D\big(\hat{q}(\cdot |w')\parallel q^*(\cdot |w)\big)
\end{eqnarray*}
were $D(\nu\parallel \mu)= \sum_{a\in A} \nu(a)\log(\nu(a)/\mu(a))$ is the Kullback-Leibler divergence between two probabilities measures $\nu$ and $\mu$ with support in same alphabet $A$.  

Now, applying successively Lemmas 6.3 and 6.2 of \cite{CsiszarTalata06},  we can upper bound the last expression above by
\begin{eqnarray*}
\sum_{w\in \tau}  \sum_{\substack{ {w'\in \tau}\\{w\prec w'} } } N^{X}_n(w')\sum_{a\in A} \frac{(\hat q(a|w')-q^*(a|w))^2}{q^*(a|w)} \leq  \sum_{w\in \tau}  \sum_{\substack{ {w'\in \tau}\\{w\prec w'} } } N^{X}_n(w')|A|\frac{1}{q^*_{min}}\frac{c \log n}{N^{X}_n(w')}.
\end{eqnarray*}
with $q^*_{min}=\min_{w\in\tau^*,a\in A}\{q^*(a|w)>0\}.$
\end{proof}

\bigskip
\begin{center}
	{\large\bf SUPPLEMENTARY MATERIAL}
\end{center}

\begin{description}
	
	\item[SeqROCTM toolbox:] A Matlab Toolbox for the analysis of Sequences of random bbjects driven by context tree models. The toolbox implements model selection methods for both sequences of random objects driven by context tree models and context tree models. It includes several others algorithms like: simulation of these kind of stochastic processes, tuning of model selection procedures, distances and dissimilarity measures for context tree models, complexity measures for context tree models, visualization of the tree structure, among others. It is written purely in Matlab language and it is self-cointained. The toolbox also contains all the examples and data used in the present paper as well as other demos. The toolbox is freely available at \url{https://github.com/noslenh/SeqROCTM-Matlab-Toolbox}.   
\end{description}

\bibliographystyle{namedmodif}
\bibliography{Bibliography}

\begin{thebibliography}{}

\bibitem[\protect\citeauthoryear{Armstrong \bgroup \em et al.\egroup
  }{2017}]{Armstrong_2017}
Armstrong, B.~C., Frost, R., and Christiansen, M.~H., The long road of
  statistical learning research: past, present and future.
\newblock {\em IPhilosophical Transactions of the Royal Society B: Biological
  Sciences}, 372(1711), 2017.

\bibitem[\protect\citeauthoryear{B\"uhlmann and Wyner}{1999}]{buhlmann99}
B\"uhlmann, P. and Wyner, A.~J., Variable length markov chains.
\newblock {\em The Annals of Statistics}, 27(2):480--513, 1999.

\bibitem[\protect\citeauthoryear{B\"{u}hlmann}{2000}]{Buhlmann2000}
B\"{u}hlmann, P., Model selection for variable length markov chains and tuning
  the context algorithm.
\newblock {\em Annals of the Institute of Statistical Mathematics},
  52:287--315, 2000.

\bibitem[\protect\citeauthoryear{B\"{u}hlmann}{2002}]{Buhlmann2002}
B\"{u}hlmann, P., Sieve bootstrap with variable-length markov chains for
  stationary categorical time series.
\newblock {\em Journal of the American Statistical Association},
  97(458):443--471, 2002.

\bibitem[\protect\citeauthoryear{Busch \bgroup \em et al.\egroup
  }{2009}]{Leonardi2009}
Busch, J.~R., Ferrari, P.~A., Flesia, A.~G., Fraiman, R., Grynberg, S.~P., and
  Leonardi, F., Testing statistical hypothesis on random trees and applications
  to the protein classification problem.
\newblock {\em The Annals of Applied Statistics}, 3(2):542--563, 2009.

\bibitem[\protect\citeauthoryear{Castro}{2006}]{Castro2006}
Castro, B.~D.
\newblock Processos estoc\'{a}sticos conduzidos por cadeias com mem\'{o}ria de
  alcance vari\'{a}vel e o jogo do goleiro, 2006.

\bibitem[\protect\citeauthoryear{Conway}{2020}]{CONWAY2020279}
Conway, C.~M., How does the brain learn environmental structure? ten core
  principles for understanding the neurocognitive mechanisms of statistical
  learning.
\newblock {\em Neuroscience and Biobehavioral Reviews}, 112:279 -- 299, 2020.

\bibitem[\protect\citeauthoryear{Csisz{\'a}r and
  Talata}{2006}]{CsiszarTalata06}
Csisz{\'a}r, I. and Talata, Z., Context tree estimation for not necessarily
  finite memory processes, via bic and mdl.
\newblock {\em IEEE Transactions on Information Theory}, 52(3):1007--1016, 3
  2006.

\bibitem[\protect\citeauthoryear{Cuesta-Albertos \bgroup \em et al.\egroup
  }{2006}]{Cuesta2006}
Cuesta-Albertos, J.~A., Fraiman, R., and Ransford, T., Random projections and
  goodness-of-fit tests in infinite-dimensional spaces.
\newblock {\em Bulletin of the Brazilian Mathematical Society, New Series},
  37(4):477--501, 2006.

\bibitem[\protect\citeauthoryear{de Lange \bgroup \em et al.\egroup
  }{2018}]{de_lange_how_2018}
Lange, F.~P.de~, Heilbron, M., and Kok, P., How {Do} {Expectations} {Shape}
  {Perception}?
\newblock {\em Trends in Cognitive Sciences}, 22(9):764--779, September 2018.

\bibitem[\protect\citeauthoryear{Duarte \bgroup \em et al.\egroup
  }{2006}]{DGG06}
Duarte, D., Galves, A., and Garcia, N.~L., Markov approximation and consistent
  estimation of unbounded probabilistic suffix trees.
\newblock {\em Bulletin of the Brazilian Mathematical Society}, 37(4):581--592,
  Dec 2006.

\bibitem[\protect\citeauthoryear{Duarte \bgroup \em et al.\egroup
  }{2019}]{Duarte:19}
Duarte, A., Fraiman, R., Galves, A., Ost, G., and Vargas, C.~D., Retrieving a
  context tree from eeg data.
\newblock {\em Mathematics}, 7(5), 2019.

\bibitem[\protect\citeauthoryear{Fern{\'a}ndez and
  Galves}{2002}]{FernandezGalves02}
Fern{\'a}ndez, R. and Galves, A., Markov approximations of chains of infinite
  order.
\newblock {\em Bulletin of the Brazilian Mathematical Society}, 33(3):295--306,
  Nov 2002.

\bibitem[\protect\citeauthoryear{Galves and
  Leonardi}{2008}]{Galves-Leonardi2008}
Galves, A. and Leonardi, F., Exponential inequalities for empirical unbounded
  context trees.
\newblock {\em In and Out of Equilibrium 2}, 2008.

\bibitem[\protect\citeauthoryear{Galves and
  L\"ocherbach}{2008}]{Galves-Loch:08}
Galves, A. and L\"ocherbach, E., Stochastic chains with memory of variable
  length.
\newblock {\em {TICSP} Series}, 38:117--133, 2008.

\bibitem[\protect\citeauthoryear{Galves \bgroup \em et al.\egroup
  }{2012}]{galves:12}
Galves, A., Galves, C., Garc{\' \i}~a, J.~E., Garcia, N.~L., and Leonardi, F.,
  Context tree selection and linguistic rhythm retrieval from written texts.
\newblock {\em Ann. Appl. Stat.}, 6(1):186--209, 2012.

\bibitem[\protect\citeauthoryear{Garrido \bgroup \em et al.\egroup
  }{2013}]{Garrido:13}
Garrido, M.~I., Sahani, M., and Dolan, R.~J., Outlier responses reflect
  sensitivity to statistical structure in the human brain.
\newblock {\em PLOS Computational Biology}, 9(3), 2013.

\bibitem[\protect\citeauthoryear{Hern\'{a}ndez \bgroup \em et al.\egroup
  }{2021}]{Hernandez:21}
Hern\'{a}ndez, N., Duarte, A., Ost, G., Fraiman, R., Galves, A., and Vargas,
  C.~D., Retrieving the structure of probabilistic sequences of auditory
  stimuli from eeg data.
\newblock {\em Scientific Reports}, 11(3520), 2021.

\bibitem[\protect\citeauthoryear{M\"{a}chler and
  B\"{u}hlmann}{2004}]{Machler2004}
M\"{a}chler, M. and B\"{u}hlmann, P., Variable length markov chains:
  Methodology, computing, and software.
\newblock {\em Journal of Computational and Graphical Statistics},
  13(2):435--455, 2004.

\bibitem[\protect\citeauthoryear{Mill}{2020}]{Mill}
Mill, B., Drawing presentable trees.
\newblock {\em Python Magazine, https://llimllib.github.io/pymag-trees/}, 2020.

\bibitem[\protect\citeauthoryear{Rissanen}{1983}]{Rissanen:83}
Rissanen, J., A universal data compression system.
\newblock {\em IEEE Trans. Inf. Theor.}, 29(5):656--664, 1983.

\bibitem[\protect\citeauthoryear{Schapiro and
  Turk-Browne}{2015}]{SCHAPIRO2015501}
Schapiro, A. and Turk-Browne, N.
\newblock Statistical learning.
\newblock In Toga, A.~W., editor, {\em Brain Mapping}, pages 501 -- 506.
  Academic Press, Waltham, 2015.

\bibitem[\protect\citeauthoryear{Stern \bgroup \em et al.\egroup
  }{2020}]{stern2020}
Stern, R.~B., d'Alencar, M.~S., Uscapi, Y.~L., Gubitoso, M.~D., Roque, A.~C.,
  Helene, A.~F., and Piemonte, M.~E.~P., Goalkeeper game: A new assessment tool
  for prediction of gait performance under complex condition in people with
  parkinson's disease.
\newblock 50(12), 2020.

\bibitem[\protect\citeauthoryear{Summerfield and de
  Lange}{2014}]{summerfield_expectation_2014}
Summerfield, C. and Lange, F.~P.de~, Expectation in perceptual decision making:
  neural and computational mechanisms.
\newblock {\em Nature Reviews Neuroscience}, 15(11):745--756, November 2014.

\bibitem[\protect\citeauthoryear{van~der Geest}{2019}]{permn}
Geest, J.van~der , Permutations with repetition, all or a subset.
\newblock {\em MATLAB Central File Exchange}, 2019.

\bibitem[\protect\citeauthoryear{von Helmholtz}{1867}]{VonHelmholtz:67}
Helmholtz, H.von .
\newblock {\em Handbuch der physiologischen Optik}, volume III.
\newblock Leopold Voss, 1867.
\newblock translated by The Optical Society of America in 1924 from the third
  germand edition, 1910, Treatise on physiological optics, Vol. III.

\bibitem[\protect\citeauthoryear{Wacongne \bgroup \em et al.\egroup
  }{2012}]{Wacongne:12}
Wacongne, C., Changeux, J., and Dehaene, S., A neuronal model of predictive
  coding accounting for the mismatch negativity.
\newblock {\em The Journal of Neuroscience}, 32(11):3665--3678, 2012.

\end{thebibliography}

\end{document}